\documentclass{article}

\usepackage[preprint]{nips_2018}

\usepackage[utf8]{inputenc} 
\usepackage[T1]{fontenc}    
\usepackage{hyperref}       
\usepackage{url}            
\usepackage{booktabs}       
\usepackage{amsfonts}       
\usepackage{nicefrac}       
\usepackage{microtype}      

\usepackage{comment}
\usepackage{graphicx} 
\usepackage{subfigure} 
\usepackage{pdflscape}
\usepackage{algorithm}
\usepackage{algorithmic}
\usepackage{bm}
\usepackage{soul}
\usepackage{listings}

\usepackage{multirow}
\usepackage{multicol}

\def\finalversion{0}

\usepackage[framemethod=tikz]{mdframed}
\usepackage{geometry}
\usepackage[many]{tcolorbox}
\usepackage{marginnote}

\newtcolorbox[use counter=mynote]
  {mynote}[1][]
  {title=ToDo~\thetcbcounter,
   width=1.3in, 
   left=0pt,
   right=0pt,
   fonttitle=\bfseries\color{black},
   colframe=pink,
   colback=pink!10,
   #1
}

\newcommand{\solicitation}[1]{\if\finalversion1 {} \else \begin{mdframed}[hidealllines=true,backgroundcolor=blue!20]From Solicitation: \textit{#1}\end{mdframed} \fi}

\newcommand\Tstrut{\rule{0pt}{2.2ex}}         

\title{
On PyTorch Implementation of Density Estimators for von Mises-Fisher and Its Mixture 
}

\author{
  Minyoung Kim\thanks{\texttt{mikim21@gmail.com}} \\
}

\begin{document}

\maketitle

\begin{abstract}
The von Mises-Fisher (vMF) is a well-known density model for directional random variables. The recent surge of the deep embedding methodologies for high-dimensional structured data such as images or texts, aimed at extracting salient directional information, can make the vMF model even more popular. In this article, we will review the vMF model and its mixture, provide detailed recipes of how to train the models, focusing on the  maximum likelihood estimators, in Python/PyTorch. In particular, implementation of vMF typically suffers from the notorious numerical issue of the Bessel function evaluation in the density normalizer, especially when the dimensionality is high, and we address the issue using the MPMath library that supports arbitrary precision. For the mixture learning, we provide both  minibatch-based large-scale SGD learning, as well as the EM algorithm which is a full batch estimator. 
For each estimator/methodology, we test our implementation on some synthetic data, while we also demonstrate the use case in a more realistic scenario of image clustering. 
Our code is publicly available in \url{https://github.com/minyoungkim21/vmf-lib}. 
\end{abstract}
\section{Definition of von Mises-Fisher Density}

The von Mises-Fisher (vMF for short) distribution, defined over the unit hypersphere in $\mathbb{R}^d$, has the following density function:
\begin{equation}
p({\bf x}; {\bm \mu}, \kappa) = C_d(\kappa) \cdot \exp(\kappa {\bm \mu}^\top {\bf x}), \ \ \ \ C_d(\kappa) = \frac{\kappa^{d/2-1}}{(2\pi)^{d/2} I_{d/2-1}(\kappa)}, 
\end{equation}
where $I_\alpha$ is the modified Bessel function of the first kind with order $\alpha$. 
Here, $\kappa$ (scalar) and ${\bm \mu}\in\mathbb{R}^d$ constitute the parameters of the vMF density, and they are constrained as: $\kappa \geq 0$ and $||{\bm \mu}|| = 1$. 
Obviously, the density has a single mode at ${\bf x} = {\bm \mu}$, and any hyperplane normal to ${\bm \mu}$ forms a level set, that is, the likelihood remains unchanged for $\{ {\bf x}: {\bm \mu}^\top {\bf x} = \gamma \}$ with $|\gamma| \leq 1$. Due to the symmetry of the density along the vector ${\bm \mu}$, the mean of the density is also ${\bm \mu}$. And, $\kappa$ determines how peaky the distribution is around its mode/mean ${\bm \mu}$, thus it is often called the {\em concentration} parameter (i.e., large $\kappa$ means that the density is more concentrated around ${\bm \mu}$, and vice versa). 
Note also that the normalizer $C_d(\kappa)$ depends only on $\kappa$, not ${\bm \mu}$, due to the symmetry of the density function.

\section{MLE for vMF}

We derive maximum likelihood estimators (MLEs) for the vMF density function, given the training data $\{ {\bf x}_i \}_{i=1}^N$. It is known that the MLE for vMF admits a {\em near} closed-form  formula, where by {\em near} one has to resort to some approximation due to the difficulty of inverting the Bessel ratio function. We will derive this in Sec.~\ref{sec:mle_vmf_batch} below by following the approximation schemes proposed in~\citep{banerjee05}, and this estimator naturally forms a {\em full-batch learner} in that we need to go through the entire data to have a model update. Alternatively we can perform the gradient ascent of the log-likelihood objective function, which is amenable to stochastic online (mini-batch) learning as discussed in Sec.~\ref{sec:mle_vmf_sgd}.

\subsection{Full-Batch Learning (Near Closed Form)}\label{sec:mle_vmf_batch}


The maximum likelihood learning can be written as the following optimization problem:
\begin{align}
&\max_{{\bm \mu}, \kappa} \ \mathcal{L}({\bm \mu}, \kappa) \ := \  \sum_{i=1}^N \log p({\bf x}_i) = N\kappa {\bm \mu}^\top \overline{{\bf x}} + N \log C_d(\kappa) \\
& \ \ \textrm{s.t.} \ \ ||\mu|| = 1 \ \  \textrm{and} \ \ \kappa \geq 0,
\end{align}
where $\overline{{\bf x}} = \frac{1}{N} \sum_{i=1}^N {\bf x}_i$. 
By taking the derivatives of the objective wrt parameters and setting them to $0$, it is easy to see that the optimum $({\bm \mu}^*, \kappa^*)$ should meet:
\begin{align}
{\bm\mu}^* = \frac{\overline{{\bf x}}}{||\overline{{\bf x}}||} \ \ \ \  \textrm{and} \ \ \ \ 
\frac{C'_d(\kappa^*)}{C_d(\kappa^*)} = -||\overline{{\bf x}}||, \label{eq:vmf_mle_eqs}
\end{align}
where 
$f'(a) = \frac{d f(a)}{d a}$. 

For the second equation in (\ref{eq:vmf_mle_eqs}), by letting $\xi = (2\pi)^{d/2}$ and $s=d/2-1$, we have:
\begin{equation}
C_d(\kappa) = \frac{\kappa^s}{\xi \cdot I_s(\kappa)} \ \ \ \ \textrm{and} \ \ \ \  C'_d(\kappa) = \frac{1}{\xi} \cdot \frac{s \kappa^{s-1} I_s(\kappa) - \kappa^s I'_s(\kappa)}{I_s(\kappa)^2}. 
\end{equation}
Then it follows that 
\begin{equation}
\frac{C'_d(\kappa)}{C_d(\kappa)} = \frac{s}{\kappa} - \frac{I'_s(\kappa)}{I_s(\kappa)}.
\end{equation}
Using the recursive formulas for the Bessel function, more specifically, 
\begin{equation}
\frac{I'_s(\kappa)}{I_s(\kappa)} = \frac{s}{\kappa} +  \frac{I'_{s+1}(\kappa)}{I_s(\kappa)},
\end{equation}
the second condition in (\ref{eq:vmf_mle_eqs}) boils down to the following equation (for $\kappa$):
\begin{equation}
\frac{I_{d/2}(\kappa)}{I_{d/2-1}(\kappa)} = ||\overline{{\bf x}}||. 
\label{eq:bessel_ratio_eq}
\end{equation}
Unfortunately, there is no known closed-form solution to the Bessel ratio inversion problem (\ref{eq:bessel_ratio_eq}). While there have been several approximate solutions to  (\ref{eq:bessel_ratio_eq})~\citep{vmf_book,tanabe2007,sra2012}, here we focus on the approximation schemes based on the continued fraction form of the Bessel ratio function, namely
\begin{equation}
R := \frac{I_{d/2}(\kappa)}{I_{d/2-1}(\kappa)} = \frac{1}{ \frac{d}{\kappa} + \frac{1}{ \frac{d+2}{\kappa} + \cdots} } \approx 
\frac{1}{\frac{d}{\kappa} + R}.
\label{eq:cont_frac}
\end{equation}
With $R = ||\overline{{\bf x}}||$, the approximation in (\ref{eq:cont_frac}) leads to $\kappa^* \approx \frac{d\cdot ||\overline{{\bf x}}||}{1-||\overline{{\bf x}}||^2}$. To reduce the approximation error further, in~\citep{banerjee05}, a correction term is added to the numerator, 
\begin{equation}
\kappa^* \approx \frac{d\cdot ||\overline{{\bf x}}|| - ||\overline{{\bf x}}||^3 } {1-||\overline{{\bf x}}||^2}.
\label{eq:kappa_mle}
\end{equation}

\subsection{Stochastic Gradient Descent (SGD) Learning}\label{sec:mle_vmf_sgd}

To circumvent the inverse Bessel ratio approximation used in the previous full-batch learning, we can consider (stochastic) gradient descent of the negative log-likelihood loss function. 
Assuming that we can sample a mini-batch $B$ ($\ni {\bf x}$) from the training data, the negative expected log-likelihood objective on $B$ can be written as:
\begin{equation}
NELL_B = \kappa {\bm \mu}^\top \overline{{\bf x}}_B + \log C_d(\kappa),
\end{equation}
where $\overline{{\bf x}}_B = \frac{1}{|B|} \sum_{{\bf x}\in B} {\bf x}$. 

Note that the gradients of the objective can be derived as:
\begin{equation}
\frac{\partial NELL_B}{\partial {\bm \mu}} = \kappa \cdot \overline{{\bf x}}_B, \ \ \ \ 
\frac{\partial NELL_B}{\partial \kappa} = {\bm \mu}^\top \overline{{\bf x}}_B - \frac{I_{s+1}(\kappa)}{I_s(\kappa)} \ \ \ \ (s = d/2-1).
\end{equation}
And, to evaluate the objective itself, one needs to compute $\log I_s(\kappa)$ since
\begin{equation}
\log C_d(\kappa) = s \log \kappa - \frac{d}{2} \log 2\pi - \log I_s(\kappa).
\end{equation}

Hence, to perform the SGD training, the key quantities that we need to compute are:  $\log I_s(\kappa)$ and $\frac{I_{s+1}(\kappa)}{I_s(\kappa)}$. To be more specific, in the PyTorch implementation with auto-differentiation capability, the former will be used in the \texttt{forward()} method definition of \texttt{autograd.Function}, while the latter placed in the \texttt{backward()} method. 

\begin{figure}
\begin{center}
\includegraphics[trim = 18mm 0mm 0mm 0mm, clip, scale=0.625
]{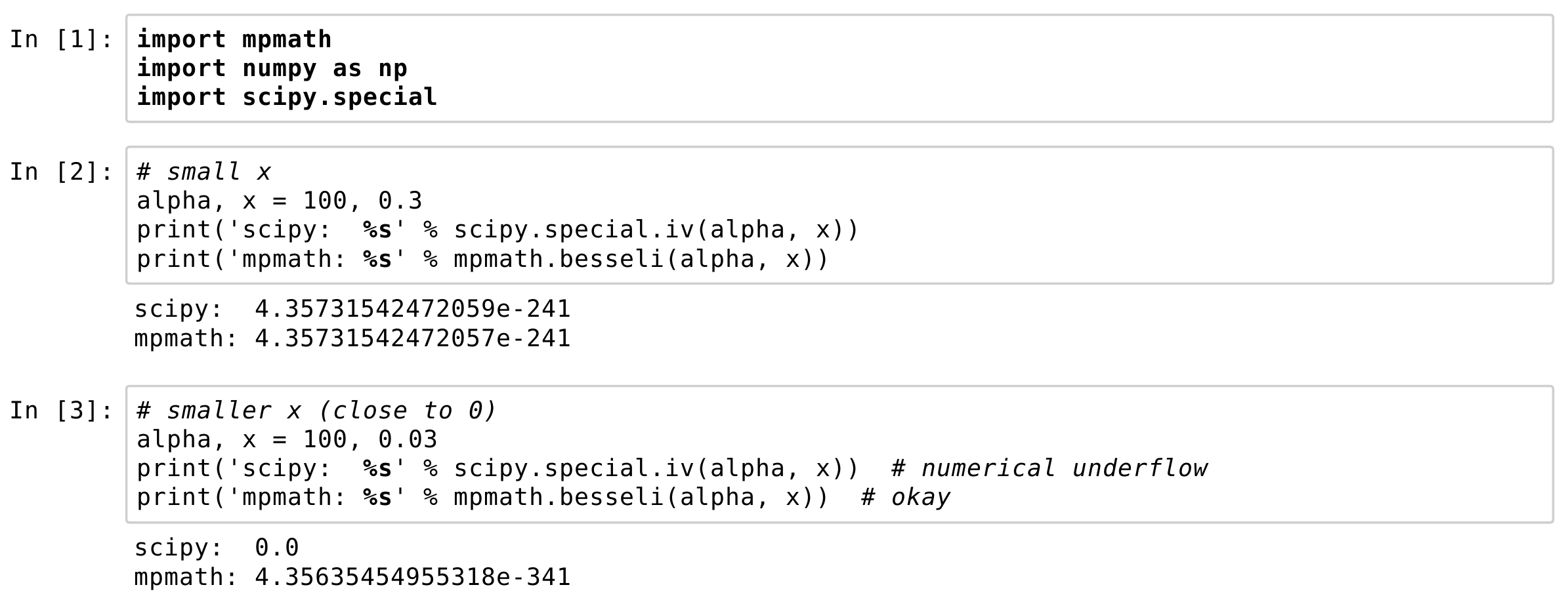} 
\end{center}
\vspace{-1.0em}
\caption{Comparison between \texttt{scipy} and \texttt{mpmath} for the Bessel function evaluation on small inputs. For the smaller input $x=0.03$, \texttt{scipy} suffers from numerical underflow. 
}
\vspace{-1.0em}
\label{fig:scipy_mpmath}
\end{figure}

Although one can use the \texttt{scipy.special.iv()} for the Bessel function of the first kind, it can easily incur numerical issues when the dimesnion $d$ becomes large (esp., when $d$ is large {\em and} $\kappa$ is close to $0$)\footnote{In the S-VAE~\citep{s-vae18}, the deep latent variable model that postulates a unit hyperspherical latent space, for instance, they introduced a vMF variational posterior, and used the \texttt{scipy.special.iv()} for the Bessel. The consequence is that the results become unstable when the dimension is greater than 50.}. 
To address the problem, we rather adopt   \texttt{mpmath}~\citep{mpmath}, the numerical Python library that supports arbitrary precision. 
The code snippet in Fig.~\ref{fig:scipy_mpmath} illustrates the basic usage of \texttt{mpmath} (e.g., $I_\alpha(x)$ can be computed by calling   \texttt{mpmath.besseli($\alpha$,x)}), while we also contrast the numerical stability between \texttt{scipy} and \texttt{mpmath} for high-order and near-$0$ input cases. 




Here is another useful tip. For the computational efficiency, the block operation might be required. That is, we need to apply \texttt{mpmath} functions to all elements in a tensor/array. For the \texttt{numpy}, this can be done by vectorizing the \texttt{mpmath} function using \texttt{numpy.vectorize()}. For instance, the following code snippet computes $I_{100}(x)$ for all elements $x$ in a $(3 \times 4)$ \texttt{numpy} array.
\begin{lstlisting}
    >>> arr = numpy.random.rand(3,4)*mpmath.mpf(1.0) 
    >>> besseli = numpy.vectorize(mpmath.besseli)
    >>> besseli(100, arr)
\end{lstlisting}

A clear advantage of defining the \texttt{autograd.Function} for the gradients of the log-likelihood of the vMF density wrt the vMF parameters, is that (using the chain rule) we can easily optimize a model that has  tensors/parameters complexly related to the vMF parameters. For instance, we may define the vMF parameters to be the outputs of some neural networks, say ${\bm\mu}_{{\bm\nu}}({\bf y})$ and/or $\kappa_{{\bm\lambda}}({\bf y})$, with weight parameters ${\bm\nu}$ and ${\bm\lambda}$ on the input ${\bf y}$.

\subsection{Empirical Comparison between Full-Batch Learning and SGD}\label{sec:batch_vs_sgd}

For this empirical comparison, we choose a vMF density $p({\bf x}; {\bm\mu}^*, \kappa^*)$, and sample iid data $\{{\bf x}_i\}_{i=1}^N$ from it. Then the true parameters $({\bm\mu}^*, \kappa^*)$ are estimated using the two estimators discussed above, with the sampled data. To sample from a vMF, we follow the scheme in~\citep{vmf_samp}. To be concrete, sampling a $d$-dim ${\bf x} \sim p({\bf x}; {\bm\mu}, \kappa)$ can be done by the following steps:
\begin{enumerate}
\item Sample ${\bf v}$, a unit vector in $\mathbb{R}^{d-1}$ uniformly. (It can be done by first sampling $d-1$ iid $\mathcal{N}(0,1)$ numbers, forming a vector, then unit-normalizing it.)
\item Sample $v_0 \sim p(v_0) \propto e^{\kappa v_0} (1-v_0^2)^{0.5(d-3)}$, e.g., by acceptance-rejection sampling. 
\item Set ${\bf v} := [ v_0; (1-v_0^2)^{\frac{1}{2}}{\bf v} ]$. 
\item Compute ${\bf U} = {\bf I} - 2 \frac{ ({\bf e}_1-{\bm\mu}) ({\bf e}_1-{\bm\mu})^\top } {||{\bf e}_1-{\bm\mu}||^2}$.
\item Return ${\bf x} := {\bf U}{\bf v}$.
\end{enumerate}

The estimation errors are shown in Table~\ref{tab:mle_for_vmf}. We report the relative errors of the estimated parameters, that is, $e_{{\bm\mu}} = \frac{||{\bm\mu} - {\bm\mu}^*||} {||{\bm\mu}^*||} = ||{\bm\mu} - {\bm\mu}^*||$ for the mean parameters, and $e_{\kappa} = \frac{|\kappa-\kappa^*|}{\kappa^*}$ for the concentration parameter, where $({\bm\mu}^*,\kappa^*)$ is the parameters of the sampling distribution. To test the estimators on diverse scenarios, we vary the dimensionality $d$ from $\{5,20,100\}$ while $\kappa^*$ is chosen to be either 50 (low concentration) or 500 (high certainty). 
We fix ${\bm\mu}^* = {\bf e}_1$, the unit vector on the first axis. 

We generate $N=10,000$ samples. For the SGD optimization, we use the Adam optimizer~\citep{adam} where the learning rate is 0.01 which decays exponentially by the rate 0.95. The batch size 128, and we run it for 100 epochs. 

\begin{table}
\centering
\caption{Comparison of the full-batch and SGD  estimators with vMF-sampled data.
}
\label{tab:mle_for_vmf}
\begin{small}
\begin{sc}
\centering
\begin{tabular}{c||c|c||c|c}
\toprule
\multirow{2}{*}{dim ($d$)} & \multicolumn{2}{c||}{$\kappa^*=50$} & \multicolumn{2}{c}{$\kappa^*=500$}
\\ \cline{2-5}
\Tstrut & Batch & SGD & Batch & SGD \\
\hline\hline
\Tstrut \multirow{2}{*}{$d=5$} & $e_{\bm\mu}=3.8 \times 10^{-6}$ & $e_{\bm\mu}=3.8 \times 10^{-6}$ & $e_{\bm\mu}=8.1 \times 10^{-7}$ & $e_{\bm\mu}=8.4 \times 10^{-7}$ \\
& $e_\kappa = 1.0 \times 10^{-2}$ & $e_\kappa = 6.2 \times 10^{-4}$ & $e_\kappa = 4.6 \times 10^{-3}$ & $e_\kappa = 3.7 \times 10^{-3}$
\\
\hline
\Tstrut \multirow{2}{*}{$d=20$} & $e_{\bm\mu}=7.4 \times 10^{-5}$ & $e_{\bm\mu}=7.4 \times 10^{-5}$ & $e_{\bm\mu}=3.4 \times 10^{-6}$ & $e_{\bm\mu}=3.3 \times 10^{-6}$ \\
& $e_\kappa = 5.4 \times 10^{-3}$ & $e_\kappa = 1.4 \times 10^{-3}$ & $e_\kappa = 2.8 \times 10^{-3}$ & $e_\kappa = 1.7 \times 10^{-3}$
\\
\hline
\Tstrut \multirow{2}{*}{$d=100$} & $e_{\bm\mu}=4.8 \times 10^{-4}$ & $e_{\bm\mu}=4.8 \times 10^{-4}$ & $e_{\bm\mu}=2.1 \times 10^{-5}$ & $e_{\bm\mu}=2.1 \times 10^{-5}$ \\
& $e_\kappa = 7.0 \times 10^{-3}$ & $e_\kappa = 5.9 \times 10^{-3}$ & $e_\kappa = 1.9 \times 10^{-3}$ & $e_\kappa = 1.1 \times 10^{-3}$
\\
\bottomrule
\end{tabular}
\end{sc}
\end{small}
\end{table}

The conclusion is as follows. The two estimators perform equally well. For the mean parameter estimation, both estimators perform nearly identically. But for the concentration parameter ($\kappa$) estimation, there are slight differences: SGD is consistently more accurate than the full-batch estimator. This may originate from the batch estimator's continued fraction approximation error. Overall, the estimation error for $\kappa$ becomes larger as the dimension $d$ increases, and as the data variability (or entropy) increases (i.e., small $\kappa^*$). 

The code to reproduce the results in this section can be found in:~\url{https://github.com/minyoungkim21/vmf-lib/mle_for_vmf.py}.

\section{MLE for Mixture of vMFs}

We consider a mixture of vMF densities. A mixture with the order (the number of components) $M$ can be written down as follows:
\begin{equation}
p({\bf x}; {\bm\theta}) = \sum_{m=1}^M \alpha_m p({\bf x}; {\bm\mu}_m, \kappa_m),  
\end{equation}
where $\alpha_m$'s are the mixing proportions. The parameters of the mixture are denoted by ${\bm\theta} = \{\alpha_m, {\bm\mu}_m, \kappa_m\}_{m=1}^M$. 
The EM algorithm~\citep{em77} is recognized as the most popular generic algorithm for mixture estimation. However, when the size of the training data is large, each E-step can be computationally demanding since one has to pass through the entire data. It might take a long time until  the model is updated by a single EM iteration. 

There are broadly two workarounds to deal with large-scale training data. The first approach is to perform the  stochastic gradient ascent on the log-marginal likelihood (i.e., taking the gradient step $\nabla_{{\bm\theta}} \frac{1}{|B|} \sum_{{\bf x}\in B} \log p({\bf x}; {\bm \theta})$ with the minibatch $B$). The second approach is the recent {\em stochastic EM}~\citep{sem,sem_var} that aims to solve a stochastic noisy version of the fixed point estimation through Robbins-Monro stochastic approximation method~\citep{robbins_monro}. Although deriving the stochastic EM for a vMF mixture can be done straightforwardly, in this section we instead dig out the former (direct stochastic gradient ascent), and empirically compare it with the full-batch EM learning.

\subsection{Derivation for EM }\label{sec:mixture_em}

The EM algorithm is essentially a block coordinate optimization method for the lower bound of the data log-likelihood, where the lower bound is obtained by applying the Jensen inequality to the log function. More specifically, letting $z \in \{1,\dots,M\}$ be the (hidden) component membership indicator, and $p_m({\bf x}) = p({\bf x}|z=m)$ be the component conditional distribution, it alternates the following two steps until convergence:
\begin{itemize}
    \item \textbf{E-step}: With ${\bm \theta}$ fixed, evaluate: $q_i(z) = p(z|{\bf x}_i; {\bm \theta})$ for  $i=1,\dots,N$.
    \item \textbf{M-step}: With $q$ fixed, solve: $\max_{{\bm\theta}} \mathcal{L}({\bm\theta}) := \sum_{i=1}^N \sum_{m=1}^M q_i(z=m) 
    \log \big( \alpha_m p_m({\bf x}_i) \big)$. 
\end{itemize}

The E-step can be done easily by
\begin{equation}
q_i(z=m) = p(z=m|{\bf x}_i)=\frac{\alpha_m p_m({\bf x}_i)}{\sum_m \alpha_m p_m({\bf x}_i)}.
\end{equation}
Note that  $p_m({\bf x}) = \textrm{vMF}({\bf x}; {\bm\mu}_m, \kappa_m)$ only requires the evaluation of the normalizer $C_d(\kappa_m)$ (and hence evaluation of the Bessel function). 
The M-step, with $q_i$'s fixed, can be derived for each parameter as follows, admitting (near) closed forms:
\begin{itemize}
\item ${\bm\mu}_m$: We set 
$\frac{\partial \mathcal{L}}{\partial {\bm\mu}_m} = 
  \sum_{i=1}^N q_i(m)   \kappa_m  {\bf x}_i$
to $0$, and solve it with the unit-norm constraint:
\begin{equation}
{\bm\mu}_m^* = \frac{\sum_{i=1}^N q_i(m) {\bf x}_i}{||\sum_{i=1}^N q_i(m) {\bf x}_i||}.
\end{equation}
\item $\kappa_m$: Setting the derivative $\sum_{i=1}^N q_i(m) \Big( {\bm \mu}_m^\top {\bf x}_i - \frac{I_{d/2}(\kappa_m)}{I_{d/2-1}(\kappa_m)} \Big)$ to $0$, leads to:
\begin{equation}
\frac{I_{d/2}(\kappa_m^*)}{I_{d/2-1}(\kappa_m^*)} = \frac{\sum_{i=1}^N q_i(m) {\bf x}_i^\top {\bm \mu}_m^* }{\sum_{i=1}^N q_i(m)} = \frac{||\sum_i q_i(m) {\bf x}_i||}{\sum_{i=1}^N q_i(m)} \ (=: R_m).
\label{eq:em_kappa_m}
\end{equation}
To solve (\ref{eq:em_kappa_m}) for $\kappa_m$, we have to resort to an approximation scheme, and we employ the continued fraction method with the correction term, as described in Sec.~\ref{sec:mle_vmf_batch}. That is, 
\begin{equation}
\kappa_m^* \approx \frac{d\cdot R_m - R_m^3 } {1-R_m^2}.
\end{equation}
\item $\alpha_m$: The maximum is attained by the empirical mean of the posteriors for the membership $m$. That is,
\begin{equation}
\alpha_m^* = \frac{\sum_{i=1}^N q_i(m)} {N}.
\end{equation}
\end{itemize}

\subsection{Derivation for Direct SGD}\label{sec:mixture_sgd}

For the large training data, we can apply the SGD by directly taking the gradient of the data log-likelihood objective over a small minibath $B$ ($\ni {\bf x}$), namely $\mathcal{L}_B := \nabla_{{\bm\theta}} \frac{1}{|B|} \sum_{{\bf x}\in B} \log p({\bf x}; {\bm \theta})$. The gradient (at a single instance ${\bf x}$) can be derived as follows: 
\begin{equation}
\nabla_{{\bm\theta}} \log p({\bf x}; {\bm\theta}) = 
  \sum_{m=1}^M p(z=m|{\bf x}) \big( \nabla_{{\bm\theta}} \log \alpha_m + \nabla_{{\bm\theta}} \log p_m({\bf x}) \big).
\end{equation}

In particular, the gradients for individual parameters can be derived as:
\begin{align}
\frac{\partial \log p({\bf x}; {\bm\theta})}{\partial {\bm\mu}_m} & \ = \
  p(z=m|{\bf x}) \kappa_m {\bf x} \\
\frac{\partial \log p({\bf x}; {\bm\theta})}{\partial \kappa_m} & \ = \
  p(z=m|{\bf x}) \bigg( {\bm \mu}_m^\top {\bf x} -    
     \frac{I_{d/2}(\kappa_m)}{I_{d/2-1}(\kappa_m)} \bigg) \\
\frac{\partial \log p({\bf x}; {\bm\theta})}{\partial \alpha_m} & \ = \ 
  p(z=m|{\bf x}) \frac{1}{\alpha_m}.
\end{align}

However, with the auto-differentiation feature provided in the PyTorch, one can bypass the implementation of the above steps: we simply form a computation graph of $\log p({\bf x}) = \log \sum_m \alpha_m p_m({\bf x})$ (e.g., using the PyTorch's \texttt{logsumexp()} function), and utilize the backprop.

\subsection{Empirical Study on Synthetic Data Clustering}\label{sec:mixture_synth}

To test the EM and SGD mixture learning algorithms, we choose a mixture model and sample iid data from it. The mixture order is set to 3 and the data dimension 5. The true parameters are: $\alpha_1 = 0.3$, $\alpha_2=0.4$, $\alpha_3=0.3$, ${\bm\mu}_1 = [0.0889, -0.3556,  0.6815,  0.1185,  0.6222]$, ${\bm\mu}_2 = [ 1.0000,  0.0000,  0.0000,  0.0000,  0.0000]$, ${\bm\mu}_3 = [-0.0889,  0.3556, -0.6815, -0.1185, -0.6222]$, and $\kappa_1=100.0$, $\kappa_2=50.0$, $\kappa_3=100.0$. We sample 1000 data points from the model.  

\begin{figure}
\begin{center}
\fbox{\includegraphics[trim = 16mm 13mm 52mm 0mm, clip, scale=0.755
]{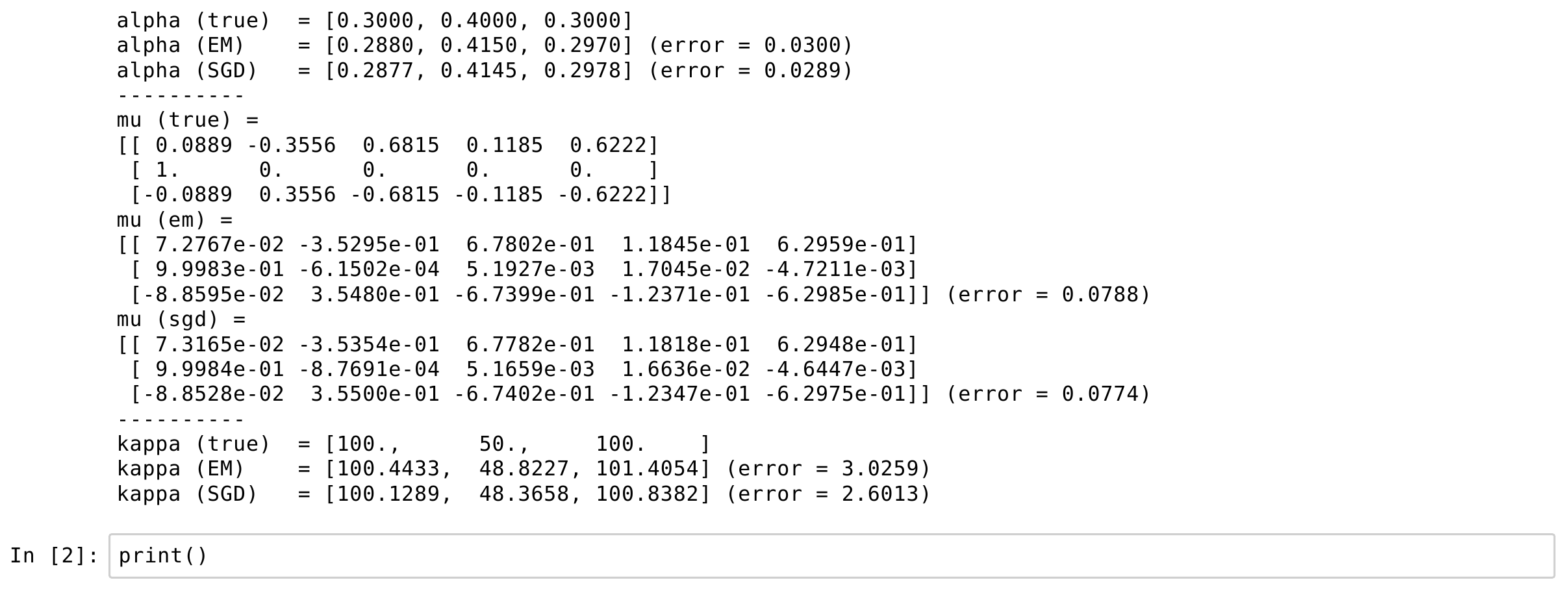}} 
\end{center}
\vspace{-0.5em}
\caption{Comparison between the (full-batch) EM and SGD learning for mixture estimation.
}
\vspace{-1.0em}
\label{fig:mixture_em_vs_sgd}
\end{figure}

For the EM algorithm we choose the maximum number of iterations 100 (but converged very quickly after a few iterations), and the tolerance of the relative log-likelihood improvement in iterations is $10^{-5}$ which serves as a stopping criterion. For the SGD learning, the batch size is set to 64, and the training goes until 100 epochs with learning rate 0.1 that decays with the rate of 0.95 every epoch. We compute the errors of the learned models, defined to be the absolute differences between true parameters and the learned parameters. To deal with the invariance of mixture components permutation in mixture models, 
we consider all permutations of the mixture components of the learned models, and take the one with the smallest error. The results are summarized in Fig.~\ref{fig:mixture_em_vs_sgd}. As shown, the two learning methods work equally well, while the SGD performs slightly better (but not significantly) in terms of the L1 error in the parameter space. 

The code to reproduce the results in this section can be found in:~\url{https://github.com/minyoungkim21/vmf-lib/mle_for_mix_vmf.py}.


\section{Application: Image Clustering in Embedded Unit-Hypersphere Space}\label{sec:apps}

In this section we test the vMF mixture model on the image clustering problem. 
%
%
In particular, we deal with image clustering with the CIFAR-10 dataset. The idea is that we first learn the useful embeddings/features ${\bf z} = e({\bf x})$ for image ${\bf x}$, which can be done by minimizing the reconstruction error in the  auto-encoding process. We impose the constraint $||{\bf z}|| = 1$ by placing a normalization layer at the end of the encoder pipeline. Once the features ${\bf z}$'s are learned, we apply the EM or SGD algorithm with the fixed ${\bf z}$ data. We choose $\dim({\bf z})=100$ for the feature dimension, and the encoder/decoder networks are built from conv/deconv layers. 

As a baseline we also compare the two algorithms with the simple K-means~\citep{kmeans} on the fixed ${\bf z}$ data. For the clustering performance metrics, we use the popular: Adjusted Rand Index (ARI) and Normalized Mutual Information (NMI), where the higher the better for both metrics. 
The results are shown in Table~\ref{tab:img_clustering}. The EM and SGD attain better performance than k-means, while both perform equally well. 

The code to reproduce the results in this section can be found in:~\url{https://github.com/minyoungkim21/vmf-lib/image_clustering_cifar10.py}.

\begin{table}
\centering
\caption{Image clustering performance on the  CIFAR-10 dataset. 
}
\label{tab:img_clustering}
\vspace{-0.5em}
\begin{small}
\begin{sc}
\centering
\begin{tabular}{c||c|c}
\toprule
\Tstrut Methods & ARI & NMI \\ \hline\hline
\Tstrut k-means & 0.0487 & 0.0871 \\ \hline
\Tstrut EM & 0.0522 & 0.1068 \\ \hline
\Tstrut SGD & 0.0528 & 0.1067 \\ 
\bottomrule
\end{tabular}
\end{sc}
\end{small}
\end{table}










\small

{
\bibliographystyle{spbasic}      
\bibliography{main}   
}

\end{document}